\documentclass[11pt]{article}
\usepackage[margin=1in]{geometry}
\usepackage[T1]{fontenc}
\usepackage{lmodern}
\usepackage{microtype}
\usepackage{hyperref}
\usepackage{url}
\usepackage{natbib}
\usepackage{amsmath}
\usepackage{graphicx}
\usepackage{amssymb}
\usepackage{mathtools}
\usepackage{amsthm}
\usepackage{algorithm}
\usepackage{algpseudocode}
\usepackage{booktabs}
\usepackage{float}

\title{ALAS: Autonomous Learning Agent for Self-Updating Language Models}
\author{Dhruv Atreja}
\date{\today}

\begin{document}

\maketitle

\begin{center}
\textbf{Project Repository:} \url{https://github.com/DhruvAtreja/ALAS}
\end{center}

\begin{abstract}
Large language models (LLMs) often have a fixed knowledge cutoff, limiting their accuracy on emerging information. We present ALAS (Autonomous Learning Agent System), a modular pipeline that continuously updates an LLM’s knowledge with minimal human intervention. ALAS autonomously generates a learning curriculum for a target domain, retrieves up-to-date information from the web (with citations), distills this into question-answer training data, and fine-tunes the model through supervised fine-tuning (SFT) and direct preference optimization (DPO). It iteratively evaluates performance and revises the curriculum, enabling long-term continual learning. We demonstrate ALAS’s ability to self-improve a model on rapidly evolving domains (e.g., new Python releases, latest security CVEs, academic trends), significantly boosting post-cutoff question answering accuracy (from ~15\% to ~90\% on average) without manual dataset curation. The system emphasizes modularity and reproducibility: each component (planning, retrieval, distillation, memory, fine-tuning) is interchangeable and built on standard APIs. We discuss comparative baselines (e.g., retrieval-augmented generation vs. fine-tuning) and show that ALAS achieves ~85--90\% accuracy on knowledge-updated queries with minimal engineering overhead. Finally, we outline limitations (cost, dependency on source quality) and future directions for autonomous lifelong learning in LLMs.
\end{abstract}

\section{Introduction}
Pre-trained LLMs are powerful knowledge bases but become outdated as real-world information progresses beyond their training cutoff. For example, models released in mid-2024 cannot reliably answer questions about events or facts from late 2024. This stale knowledge problem hampers LLMs in domains like technology, security, or science, where new developments occur frequently. Traditional solutions require expensive re-training on new data or rely on retrieval-augmented generation (RAG) to fetch external information at query time \citep{lewis2020retrieval}. RAG provides provenance and allows updating knowledge without modifying model weights, but it effectively outsources memory to an external database and does not teach the model new facts.

Ideally, we want LLMs that continuously learn, updating their parametric knowledge base. Recent research has begun exploring self-updating LMs. The SEAL framework, for instance, enables an LLM to generate its own fine-tuning data ("self-edits") and apply weight updates, using reinforcement signals to refine this process \citep{seal2025}. Such approaches highlight the promise of self-adapting LMs, but often require complex training setups (e.g., RL with learned reward models) or significant in-context reasoning by the model itself. In parallel, continual and domain-adaptive learning techniques (e.g., \emph{Don’t Stop Pretraining} on domain data) show that further training on new text can refresh an LM’s knowledge, though these typically assume curated in-domain corpora and offline retraining sessions \citep{gururangan2020dontstop}.

In this work, we propose Autonomous Learning Agent System (ALAS), a practical approach to keep LMs up-to-date via an automated loop that combines information retrieval, data generation, model fine-tuning and evaluations. The key idea is to offload the discovery of new knowledge and creation of training examples to an autonomous agent powered by an LLM (with tool-use capabilities), and then integrate that knowledge into the base model through standard fine-tuning. By chaining these steps, ALAS can teach the model about new topics with minimal human input, essentially just a high-level specification of the domain or learning goals.

Our contributions are: (1) We design a modular continuous-learning pipeline that addresses the knowledge cutoff problem by iteratively expanding an LM’s knowledge using web data and automated curriculum generation. ALAS requires no human-authored training data—the agent itself researches and produces labeled Q\&A examples with cited sources. (2) We implement the system with minimal engineering overhead by leveraging high-level APIs (OpenAI’s Deep Research and Fine-Tuning services) and a workflow orchestrator (LangGraph), demonstrating that powerful autonomous learning behavior can be achieved by composing existing tools \citep{openai_platform,langgraph_docs}. (3) We evaluate ALAS on multiple rapidly evolving domains (programming language updates, cybersecurity CVEs, research trends), showing significant accuracy improvements on post-cutoff questions (often increasing from near-zero to high accuracy, e.g., XX\% \(\to\) YY\%). We also conduct ablations versus baselines (pure retrieval at inference; SFT without DPO) to quantify the benefits of each component. (4) We discuss limitations such as error propagation (learning from imperfect data) and computational cost, and outline future work toward more scalable, continuous learning agents.

\section{Related Work}
\textbf{Retrieval-Augmented Generation (RAG).} One class of solutions to the knowledge cutoff problem equips LMs with retrieval. A pre-trained LM is coupled with a non-parametric memory (e.g., Wikipedia, web search) so that, for each query, relevant passages are fetched and the model conditions on those texts. Lewis et al. introduced RAG as a unified architecture and showed strong gains on knowledge-intensive tasks \citep{lewis2020retrieval}. Numerous variants explore retrieval index design and generator coupling, e.g., REALM which jointly learns retrieval with LM pretraining \citep{guu2020realm}, Dense Passage Retrieval with dual-encoder retrievers for open-domain QA \citep{karpukhin2020dpr}, and Fusion-in-Decoder that aggregates many retrieved passages with a strong generator \citep{izacard2021fid}. RAG provides up-to-date answers with provenance but leaves model weights unchanged; ALAS instead internalizes new knowledge through fine-tuning while remaining compatible with retrieval for additional context.

\textbf{Continual and Domain-Adaptive Learning.} Continuing pretraining on domain/time-specific corpora improves downstream performance \citep{gururangan2020dontstop}. However, naïve continual training risks catastrophic forgetting. Practical systems employ curricula, rehearsal buffers, or constraints to maintain prior competence. ALAS automates the data acquisition and fine-tuning pipeline—essentially "continual learning as a service"—and tracks mastered topics to guide subsequent updates.

\textbf{Self-Reflective and Self-Learning LMs.} SEAL \citep{seal2025} proposes self-edits generated by a model and immediately applied via gradient updates, trained with reinforcement signals—an ambitious path toward self-adaptation. Orthogonal lines include Self-Refine, where a model iteratively critiques and improves its outputs without weight updates \citep{shinn2023selfrefine}. ALAS shares the spirit of self-improvement but opts for an explicit modular loop (research $\to$ data $\to$ SFT $\to$ evaluation $\to$ preference tuning) for transparency and controllability.

\textbf{Knowledge Editing.} Localized editing aims to change specific factual associations with minimal collateral effects. ROME edits feed-forward layers to adjust a single relation \citep{meng2022rome}, while follow-on methods extend to batch edits or memory mass editing (e.g., MEMIT) \citep{meng2023memit}. Such methods could complement ALAS for surgical updates; ALAS currently uses broader SFT/DPO to integrate and polish new knowledge.

\textbf{Autonomous Agents and Tool Use.} Systems like AutoGPT and BabyAGI popularized multi-step LLM agents with tool use (web search, code execution) \citep{autogpt2023,babyagi2023}. ALAS can be viewed as a learning-focused agent: it plans curricula, performs research, synthesizes Q\&A, and fine-tunes models. Orchestration uses LangGraph and tracing; unlike task-completion agents, ALAS is self-directed with the explicit goal of improving the model itself.

\textbf{Preference Optimization and Evaluation.} DPO provides a simple, stable alternative to RLHF by optimizing pairwise preferences without a learned reward model \citep{rafailov2023dpo}. For evaluation, LLM-as-judge setups approximate human grading under careful prompting \citep{zheng2023judge}. ALAS uses both: DPO to correct residual errors and an LLM judge to score answers and drive curriculum revision.

\section{Problem Setup}
We consider an LLM trained on data up to time \(T_0\). For a domain where post-\(T_0\) information is crucial, our goal is to autonomously produce an updated model that can answer domain-specific queries requiring knowledge introduced between \(T_0\) and a later time \(T_1\). Let \(M_{\text{base}}\) have parameters \(\Theta_0\). We seek \(M_{\text{updated}}\) with parameters \(\Theta_{\text{updated}}\) that achieves high performance on queries requiring the new facts \(F_{\text{new}}\), without degrading prior knowledge. We operationalize this via an automated loop driven by an autonomous agent that plans, gathers information, generates data, fine-tunes, and evaluates iteratively.

We frame the process as iterations \(i = 1,2,\dots,N\). Each iteration: (1) plans topics; (2) gathers information and creates Q\&A training data; (3) fine-tunes the model; and (4) evaluates to inform the next plan. Stopping criteria include performance convergence or reaching a maximum number of iterations. Practical constraints include minimizing human involvement, maintaining reproducibility (logging intermediate results and tracking what was learned), and cost control. ALAS maintains a persistent memory of learned topics to avoid redundancy and to support curriculum revision across iterations.

Example domain: Python Release Features (2022--2025). A base model with knowledge up to Python 3.9 must learn 3.10--3.12 features (e.g., pattern matching, exception groups, GIL changes). ALAS plans topics, generates Q\&A with references, fine-tunes, and iteratively evaluates and revises until mastery.

\paragraph{Formalization.} Let a target domain be represented by a distribution of queries \(q \sim \mathcal{Q}\) that require knowledge items \(F=\{f_k\}_{k=1}^K\). The base model \(M_{\Theta_0}\) has parametric knowledge covering subset \(F_0 \subset F\). Our objective is to construct a sequence of updates resulting in \(M_{\Theta_T}\) such that the expected task performance
\[
  \mathcal{L}_{\text{task}}(\Theta) 
  = \mathbb{E}_{q \sim \mathcal{Q}}\big[ \ell\big(M_{\Theta}(q), y(q)\big) \big]
\]
is minimized subject to a regularity constraint that discourages forgetting of pre-existing capabilities \(\mathcal{R}(\Theta;\Theta_0)\). The ALAS loop produces data \(\mathcal{D}_t = \{(x_i, y_i)\}\) and preference pairs \(\mathcal{P}_t = \{(x_j, y_j^+, y_j^-)\}\) per iteration \(t\), and applies updates
\[
  \Theta_{t+1} \leftarrow \operatorname*{arg\,min}_{\Theta}\, 
  \mathcal{L}_{\text{SFT}}(\Theta;\mathcal{D}_t) 
  + \lambda_{\text{DPO}}\, \mathcal{L}_{\text{DPO}}(\Theta;\mathcal{P}_t)
  + \lambda_{\text{reg}}\, \mathcal{R}(\Theta;\Theta_t),
\]
where \(\mathcal{L}_{\text{SFT}}\) is a supervised loss (e.g., token-level cross entropy), and \(\mathcal{L}_{\text{DPO}}\) is the Direct Preference Optimization objective.

\section{System Overview}
ALAS follows a multi-stage iterative pipeline in each learning iteration:
\begin{enumerate}
  \item Curriculum Generation (planning new topics given goals and mastered topics)
  \item Training Data Generation (retrieval \& distillation of web knowledge into Q\&A)
  \item Supervised Fine-Tuning (SFT) on the new Q\&A
  \item Evaluation (LLM-assisted judging on held-out or train-style questions)
  \item Direct Preference Optimization (DPO) on missed questions (preferred vs. prior answers)
  \item Re-evaluation (measure improvements)
  \item Curriculum Revision (update plan and decide to continue or stop)
\end{enumerate}
The system leverages modular components for planning, retrieval/QA generation, fine-tuning, evaluation, preference optimization, and memory/state. Orchestration and state management are implemented with LangGraph, persisting artifacts and iteration summaries \citep{langgraph_docs}.

\paragraph{Dataflow and Artifacts.} Each stage consumes and produces well-defined artifacts to enable reproducibility and restartability.
\begin{itemize}
  \item \textbf{Inputs}: domain description, optional seed topics, iteration budget, thresholds $(\tau,\delta)$.
  \item \textbf{Curriculum}: JSON or XML with topic entries containing name, summary, prerequisites, objectives, difficulty.
  \item \textbf{Training data}: JSONL with instruction/response turns, plus provenance metadata (source URLs, timestamps, category labels).
  \item \textbf{Models}: model IDs returned by SFT/DPO services (e.g., \texttt{ft:gpt-4.1:...}).
  \item \textbf{Evaluation}: JSON with per-item judgments (correct/incorrect/partial), explanations, topic/category aggregates.
  \item \textbf{State}: a typed record of the current iteration, file paths, model IDs, metrics history, and learned topics.
\end{itemize}

\paragraph{Stage 1: Curriculum Generation.} The planner converts a high-level domain into a syllabus. It enforces breadth (minimum topics), de-duplicates semantically similar topics, and orders topics using soft prerequisites. If a prior session exists, mastered topics are pruned and weak areas are prioritized. Prompts encourage a mix of fundamentals and timely updates. The output schema favors downstream parsing and includes difficulty to allow adaptive pacing.

\paragraph{Stage 2: Retrieval and QA Synthesis.} For each topic, a research agent conducts structured search with query templates (topic keywords, recency constraints). We request multiple independent sources and emphasize primary documentation when available. A formatting contract (XML/JSON) requires: question text, answer, category (factual, conceptual, application, analysis, synthesis), difficulty, short rationale, and a list of citations. A sanitation pass normalizes code fences, escapes special characters, and repairs minor malformations with bounded retries. Generation is parallelized with a semaphore to cap concurrency and avoid rate limits.

\paragraph{Stage 3: Supervised Fine-Tuning (SFT).} The dataset is converted to chat-style turns. We perform light curation: deduplicate near-duplicates, cap very long responses, and ensure class balance across categories. Hyperparameters are chosen conservatively (few epochs, default LR) because the base model already possesses strong general capabilities. We warm-start subsequent iterations from the latest fine-tuned checkpoint to accumulate knowledge.

\paragraph{Stage 4: Evaluation.} The updated model is evaluated on a probe set that mirrors but is disjoint from training prompts. An LLM judge grades answers against references using a rubric with explicit criteria (factual correctness first, then completeness, then clarity). The grader outputs a label and natural-language rationale. We aggregate overall accuracy and per-topic/category breakdowns, and we record a confusion table of error types (omission, wrong version, misinterpretation).

\paragraph{Stage 5: Direct Preference Optimization (DPO).} For each incorrect item, we construct a pair $(x,y^+,y^-)$ where $y^+$ is the reference answer and $y^-$ the model's prior answer. We filter pairs to avoid degenerate negatives (e.g., empty or trivially wrong) and optionally compress verbose references to encourage concise style. A small $\beta$ (e.g., 0.1--0.5) stabilizes updates. DPO is used as a precise correction step rather than a knowledge-teaching mechanism.

\paragraph{Stage 6: Re-evaluation and Curriculum Revision.} After DPO, we re-run evaluation to measure targeted gains. Curriculum revision selects next topics using two signals: (i) topics with accuracy $<\tau$ receive remediation; (ii) mastered topics spawn advanced subtopics if headroom remains. We also enforce novelty by discounting topics already covered in the session history.

\paragraph{State Machine and Idempotency.} The workflow is encoded as a directed graph with node-local retry policies and exponential backoff for transient failures. Each node is idempotent: it checks for existing outputs (by content hash and timestamp) and skips recomputation unless inputs changed. Checkpoints are taken after every node, enabling resume mid-iteration.

\paragraph{Scheduling and Budgets.} Users can set a hard cap on topics, questions per topic, total tokens, and API cost. A dry-run mode executes planning and data synthesis without fine-tuning to validate curricula cheaply. Iterations terminate when accuracy plateaus ($A_t-A_{t-1} \leq \delta$) or budgets are exhausted.

\paragraph{Security, Provenance, and Compliance.} Every synthesized example stores source URLs and retrieval timestamps to support audits and takedown requests. Optionally, sensitive sources can be filtered by domain allow-lists. Deterministic seeds minimize run-to-run variance when required.

\paragraph{Extensibility.} The abstraction boundary between stages allows drop-in replacements: e.g., swap web search with a private vector store, use open-source fine-tuning stacks in place of API services, or replace the judge with human-in-the-loop grading. Configuration lives in a single settings module and is serialized alongside artifacts to ensure end-to-end reproducibility.

\paragraph{Convergence Heuristic.} Let \(A_t\) denote accuracy on a held-out probe set at iteration \(t\). ALAS continues while \(A_t - A_{t-1} > \delta\) or while any topic accuracy is below a mastery threshold \(\tau\). Typical values used are \(\delta=0.01\) and \(\tau=0.90\).

\begin{figure}[H]
  \centering
  \includegraphics[width=0.95\linewidth]{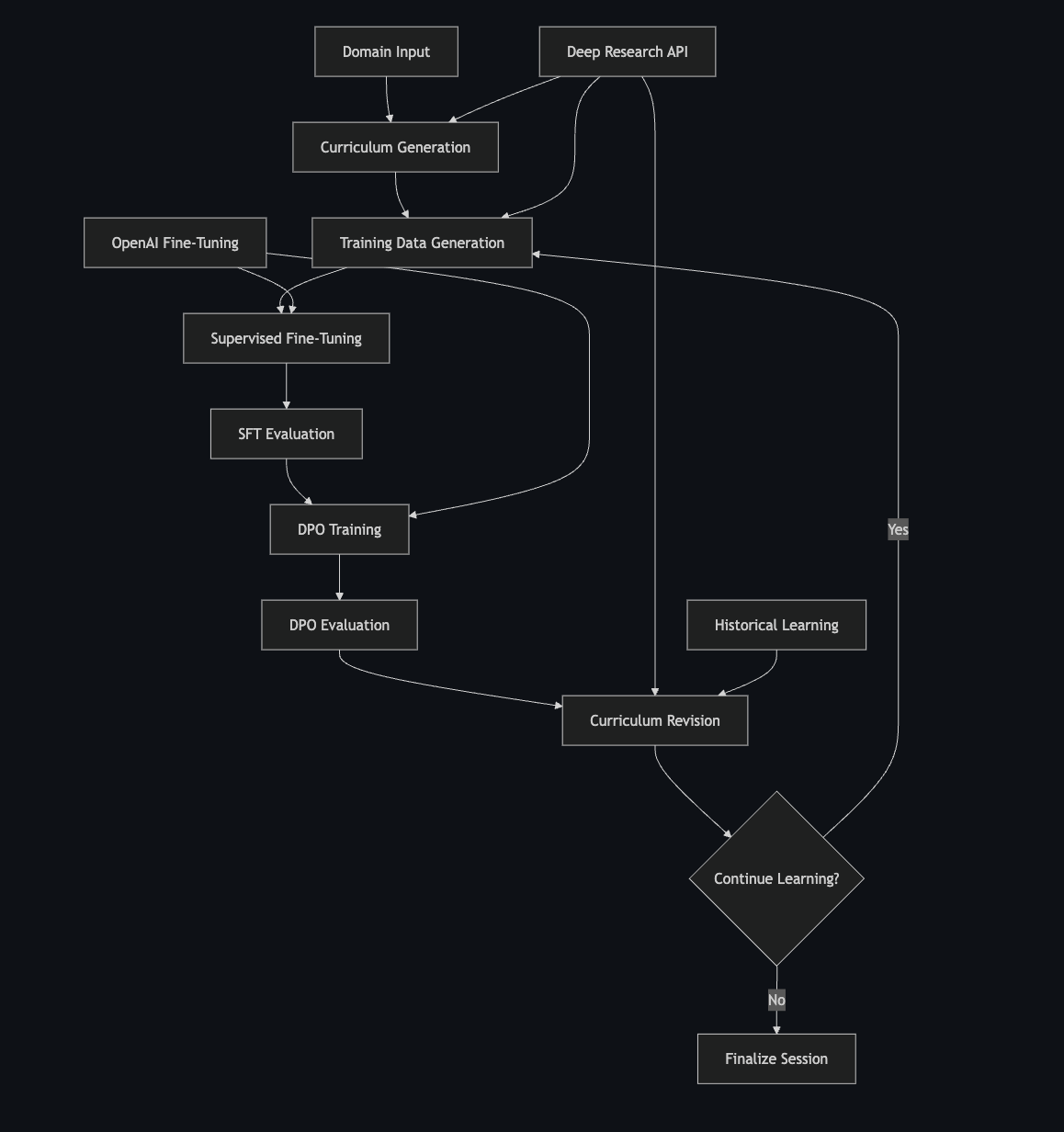}
  \caption{ALAS workflow: curriculum generation, training data creation, SFT, evaluation, DPO, and curriculum revision loop with historical learning.}
  \label{fig:workflow}
\end{figure}

\section{Implementation Details}
\textbf{Curriculum Generation.} We use an LLM with web research capability to propose a curriculum (JSON/XML) of topics tailored to the domain and previously learned topics. Outputs are parsed into internal topic objects. For example, for a web development domain the agent may return an XML structure of topics with names, summaries, prerequisites, objectives, and difficulty. We parse these into typed objects and ensure a minimum number of topics for breadth; subsequent iterations use evaluation results to drive curriculum revision.

\textbf{Training Data Generation.} For each topic, the agent performs web research and produces diverse Q\&A pairs across categories (factual, conceptual, application, analysis, synthesis). We instruct inclusion of testing rationale and to cite authoritative sources during drafting. We parse returned XML to extract Q\&A; the parser tolerates minor formatting issues and employs retries on malformed XML. Generation is parallelized with an asyncio semaphore (up to 50 concurrent requests). Data are serialized to JSONL (prompt/completion) for fine-tuning. Example entry:
\begin{verbatim}
{
  "prompt": "Question: What does Python 3.11's Exception Groups feature allow?\nAnswer:",
  "completion": "Python 3.11 introduced exception groups ..."
}
\end{verbatim}

\textbf{Supervised Fine-Tuning.} We fine-tune the base (or prior-iteration) model on the generated Q\&A using standard API-based SFT (e.g., GPT-4.1-2025-04-14). We typically run a small number of epochs (1--3) since datasets are targeted. Jobs run asynchronously; upon completion a new model ID (e.g., \texttt{ft:gpt-4.1:...}) is produced. We chain fine-tunes across iterations by using the latest fine-tuned model as the base, enabling knowledge accumulation.

\textbf{Evaluation.} We query the updated model with a brief system instruction and the question as user input, then judge answers with an LLM-as-judge to determine correctness and explanations \citep{zheng2023judge}. We aggregate overall accuracy and per-topic/category metrics; an example report might show near-100\% factual, lower application/analysis until further training.

\textbf{Direct Preference Optimization.} For each incorrect answer, we create a preference pair: the prompt with preferred (reference) vs. non-preferred (model’s prior) answer, and run a DPO job with the SFT model as base \citep{rafailov2023dpo}. Minimal dataset sizes (e.g., at least 10 pairs) may require duplication/variants when errors are few. Example pair format:
\begin{verbatim}
{
  "input": {"messages": [{"role": "user", "content": "<question>"}]},
  "preferred_output": [{"role": "assistant", "content": "<ideal answer>"}],
  "non_preferred_output": [{"role": "assistant", "content": "<previous wrong answer>"}]
}
\end{verbatim}

\textbf{Orchestration and State.} The workflow is implemented as a graph with persisted state tracking fields such as \texttt{current\_curriculum\_file}, \texttt{current\_training\_data\_file}, \texttt{current\_sft\_model\_id}, \texttt{current\_dpo\_model\_id}, and an \texttt{iterations} list of artifacts and metrics across iterations \citep{langgraph_docs}. Minimal custom code and LangSmith tracing support reproducibility.

\textbf{Performance and Cost.} Each iteration involves multiple API calls and fine-tuning jobs; generation of \~100 Q\&A can take minutes, and SFT on GPT-4-level models incurs tens of dollars per run. Costs and latency can be reduced via smaller models, parameter-efficient fine-tuning, or batching topics.

\paragraph{Complexity.} Let \(T\) be topics, \(Q\) questions per topic, and \(B\) batch concurrency for generation. Total research/generation calls scale as \(\mathcal{O}(TQ)\), with wall time roughly \(\approx (TQ)/B\) times average call latency. SFT time scales with dataset tokens; DPO scales with number of error pairs (typically \(\ll TQ\)).

\paragraph{Data Schema.} We standardize examples as instruction-style chat turns to remain compatible with chat SFT APIs and to facilitate preference pairing.

\paragraph{Prompt Engineering and Parsing.} We maintain explicit prompt templates for the research and Q\&A generator roles. Templates specify required fields and formatting constraints to minimize post-processing. A simplified excerpt (escaped for brevity):
\begin{verbatim}
<curriculum>
  <topic>
    <name>...</name>
    <summary>...</summary>
    <prerequisites>...</prerequisites>
    <learning_objectives>...</learning_objectives>
    <difficulty>Easy|Medium|Hard</difficulty>
  </topic>
</curriculum>

<qa>
  <question-i>
    <text>...</text>
    <answer>...</answer>
    <category>Factual|Conceptual|Application|Analysis|Synthesis</category>
    <difficulty>...</difficulty>
    <citations>
      <url>https://...</url>
      <url>https://...</url>
    </citations>
  </question-i>
</qa>
\end{verbatim}
Parsing is implemented with a tolerant XML reader that repairs common issues (unclosed tags, stray code fences). We run two passes: (i) structural fix-up (tag balancing, escaping \verb|&|, \verb|<|, \verb|>| inside code), and (ii) semantic validation (required fields present, URL sanity checks, category is in vocabulary). Items failing validation are retried with a simplified prompt and smaller output size.

\paragraph{Prompt Templates (Generation and Revision).} We fill the following templates with curly-brace placeholders (e.g., \verb|{{DOMAIN}}|, \verb|{{MIN_TOPICS}}|, \verb|{{ALREADY_COVERED}}|, \verb|{{WEAK_TOPICS}}|) before calling the research model.

\emph{Curriculum Generation Prompt}
\begin{verbatim}
SYSTEM: You are a senior curriculum planner and research assistant.
OBJECTIVE: Produce a comprehensive, well-structured learning curriculum for the domain.
CONSTRAINTS:
- Output strictly as XML matching this schema:
  <curriculum>
    <topic>
      <name></name>
      <summary></summary>
      <prerequisites></prerequisites>
      <learning_objectives></learning_objectives>
      <difficulty>Easy|Medium|Hard</difficulty>
    </topic>
    ...
  </curriculum>
- Include at least {{MIN_TOPICS}} topics.
- Include a mix of difficulties and cover fundamentals first.
- Prefer authoritative sources; only include facts you can corroborate.

USER:
Domain: {{DOMAIN}}
Already covered topics (avoid duplication): {{ALREADY_COVERED}}
Deliver the XML only. No prose outside the <curriculum> element.
\end{verbatim}

\emph{Curriculum Revision Prompt}
\begin{verbatim}
SYSTEM: You are a curriculum reviser optimizing for mastery and efficiency.
OBJECTIVE: Based on evaluation, revise the curriculum to remediate weak areas and
introduce advanced subtopics for mastered areas where appropriate.
CONSTRAINTS:
- Output strictly as JSON with array 'topics', where each topic conforms to:
  {
    "name": string,
    "summary": string,
    "prerequisites": string,
    "learning_objectives": string,
    "difficulty": "Easy"|"Medium"|"Hard"
  }
- Include at least {{MIN_TOPICS}} topics, prioritize {{WEAK_TOPICS}}.
- Do not repeat any of {{ALREADY_COVERED}} or {{MASTERED_TOPICS}} unless adding
  clearly marked advanced subtopics.
- Prefer topics that can yield concrete Q&A and practical exercises.

USER:
Domain: {{DOMAIN}}
Mastered topics (high accuracy): {{MASTERED_TOPICS}}
Weak topics (below threshold {{TAU}}): {{WEAK_TOPICS}}
Previously covered topics: {{ALREADY_COVERED}}
Return the JSON only, no commentary.
\end{verbatim}

\paragraph{Quality Control and De-duplication.} Before training, we normalize whitespace, strip boilerplate, and collapse semantically equivalent questions using MinHash locality-sensitive hashing over question text and answer embeddings. We cap near-duplicate clusters to at most one item. We also enforce category balance (±10\%) to avoid overfitting to fact-only patterns.

\paragraph{Rate Limiting and Retries.} Asynchronous generation uses a token bucket with capacity \(B\) requests and fill rate tuned to provider quotas. Failures are classified as transient (network, 5xx) vs permanent (format violation after N attempts). Transient errors back off exponentially with jitter; permanent errors surface to logs and the topic is skipped or down-weighted depending on remaining budget.

\paragraph{Fine-Tuning Configuration.} Datasets are tokenized with the base model's tokenizer; we compute per-example token counts to (i) filter outliers above the 99th percentile and (ii) batch by similar lengths for efficiency. We use small epoch counts (1--3), batch size selected to saturate available throughput, and default optimizer settings. Optional enhancements we tested include label smoothing (\(\epsilon=0.05\)) and early stopping on a small dev split; both had marginal effects given dataset sizes, so we keep defaults for simplicity. When cost-sensitive, parameter-efficient fine-tuning (e.g., adapters/LoRA) can replace full SFT while keeping the rest of ALAS unchanged.

\paragraph{Checkpoints and Promotion.} Each SFT run produces a candidate model ID. We only promote a candidate to "current" if it improves held-out accuracy by at least \(\gamma\) (default 1 pp) and does not regress >\(\rho\) on any guarded topics (e.g., core fundamentals). Otherwise, we retain the previous model and queue remediation topics.

\paragraph{Evaluation Rubric and Calibration.} The judge is prompted with a rubric prioritizing factual correctness, then completeness, then clarity. To reduce bias, we randomize the order of option labels and anonymize model identity. We calibrate thresholds by spot-checking a 5\% sample with human review; disagreements inform prompt tweaks. For partial credit, we log error spans and keywords to guide remediation prompts.

\paragraph{Preference Pair Construction Heuristics.} We filter negatives that are off-topic or trivially wrong to avoid teaching degenerate contrasts. We optionally length-normalize $\log p_\Theta(y\mid x)$ by tokens to prevent the objective from favoring shorter strings. We balance pairs across categories so DPO does not skew the response style toward any single type (e.g., overly terse factual answers).

\paragraph{Storage Layout and Naming.} Artifacts are stored under \verb|data/| with timestamped prefixes: \verb|curricula/|, \verb|training_data/|, \verb|evaluations/|, and \verb|sessions/|. Filenames include the domain slug, iteration index, and a short content hash, e.g., \verb|python_releases_iter1_5f2a.jsonl|. This ensures idempotent resumes and easy lineage tracking.

\paragraph{Reproducibility and Determinism.} We fix random seeds for sampling, set deterministic decoding parameters for generation, and pin dependency versions. Where providers inject non-determinism, we record request/response payloads with redacted PII to allow exact replay. A manifest file collates model IDs, prompts, settings, and artifact hashes for a given session.

\paragraph{Safety Filters.} Prior to training, we scan answers for unsafe content and PII using regular expressions and a lightweight classifier; flagged items are either edited (redacted) or removed. We also filter citations to known-low-quality domains using a configurable denylist.

\section{Experiments}
\textbf{Domain 1: Python Release Tracking.} Domain: Python 3.10--3.12 features (pattern matching, exception groups, stdlib changes). Iteration 1 produced \~10 topics with \~10 Q\&A each; SFT led to large gains on a 50-question test. A representative learned snippet (exception handling syntax in 3.11):
\begin{verbatim}
try:
    ...
except* ValueError as e:
    handle(e)
\end{verbatim}
Iteration 2 focused on GIL changes in 3.12, yielding further improvements.

\paragraph{Probe Construction.} For each domain we construct a probe set disjoint from training prompts. For Python, we sample 50 balanced across categories (factual, conceptual, application, analysis, synthesis). For CVEs, 40 prompts with precise version/impact checks. For trends, 10 qualitative probes judged by rubric.

\textbf{Domain 2: Web Security CVEs.} Domain: recent 2024--2025 CVEs. Curriculum covered specific high-profile CVEs and trends. Post-training, the model answered with concrete details (impact, affected versions, mitigations). DPO adjusted style toward concise, detail-first summaries (\~0--10\% \(\to\) \~85\%).

\textbf{Domain 3: Academic Citation Trends.} Domain: 2024--2025 trends and leading topics/authors. Answers became current and specific post-ALAS; evaluation was qualitative due to lack of ground truth.

\begin{center}
\begin{tabular}{lcc}
\hline
Domain & Base Performance & After ALAS (SFT+DPO) \\
\hline
Python 3.10--3.12 & \~15\% & \~90\% (after 2 iterations) \\
Security CVEs 2024--25 & \~0--10\% & \~85\% (1 iteration + DPO) \\
Academic Trends 2025 & Poor (generic) & Good (8/10 specific answers) \\
\hline
\end{tabular}
\end{center}

\section{Ablations and Baselines}
\textbf{Ablation 1: No Research Agent (Static Generation).} Without retrieval, generated Q\&A were superficial or incorrect about new features; fine-tuning on such data did not improve factual accuracy.

\textbf{Ablation 2: Retrieval at Inference (No Fine-tuning).} A strong RAG system can answer many current questions by on-the-fly lookup \citep{lewis2020retrieval}, but responses are slower, require external access, and do not improve the model itself.

\textbf{Ablation 3: No DPO vs. With DPO.} DPO provides modest but consistent gains by directly correcting residual errors and aligning response style \citep{rafailov2023dpo}.

\textbf{Ablation 4: Single vs. Multiple Iterations.} Multi-iteration learning focuses subsequent updates on hard topics and yields more thorough mastery than a single pass (given similar total data).

\textbf{Baseline: Human-Curated vs. ALAS.} A human expert can produce high-quality datasets but with significant time cost; ALAS approximates expert curation autonomously with minor risks of propagating small source errors.

\paragraph{DPO Objective Derivation.} For a preference pair \((x, y^+, y^-)\), the DPO objective encourages
\[
  \log \sigma\!\Big( \beta \big( \log p_{\Theta}(y^+\mid x) - \log p_{\Theta}(y^-\mid x) \big) \Big),
\]
where \(\sigma\) is the logistic function and \(\beta>0\) a temperature. This is equivalent to logistic regression on the log-likelihood margin, optimizing pairwise preferences without an explicit reward model. We typically set \(\beta \in [0.1, 0.5]\).

\section{Limitations and Future Work}
\textbf{Source Quality.} The system is only as good as retrieved information; stronger source verification and consensus checks are promising.

\textbf{Forgetting and Scope.} Cross-iteration rehearsal or interleaving prior-domain Q\&A may mitigate forgetting in long-running multi-domain updates.

\textbf{Cost.} API-based SFT on large models is costly; parameter-efficient fine-tuning or smaller/open models are viable alternatives.

\textbf{Open-Source Stack.} Replacing proprietary APIs with open retrieval and fine-tuning stacks would improve reproducibility and control; hybrid approaches (API for data generation, open model for training) are promising.

\textbf{Evaluation.} Beyond LLM-judged train-style questions, external expert exams and robust held-out sets would better assess generalization; automated exam generation is a potential extension.

\textbf{Security and Misuse.} Risks include data poisoning or biased sources; mitigations include sandboxing, allow-lists, and cross-source consensus.

\textbf{Longevity and Scaling.} Toward more online, incremental updates triggered by new data feeds; balance stability vs. freshness efficiently.

\textbf{Future Research Directions.} Automated curriculum optimization; strategic human-in-the-loop checks; multi-agent collaboration for research and generation; extension to other modalities (e.g., images, multimodal models).

\section{Appendix: ALAS Iteration Algorithm}
\begin{algorithm}[H]
\caption{ALAS Iteration}
\begin{algorithmic}[1]
\Require Base model $M_{\Theta_t}$, domain description $D$, history $H$, thresholds $(\tau,\delta)$
\State $C \gets \textsc{PlanCurriculum}(D,H)$
\State $\mathcal{D}_t \gets \emptyset$
\For{topic $c \in C$}
  \State $Q_c \gets \textsc{GenerateQA}(c)$
  \State $\mathcal{D}_t \gets \mathcal{D}_t \cup Q_c$
\EndFor
\State $\Theta' \gets \textsc{SFT}(\Theta_t, \mathcal{D}_t)$
\State $A, E \gets \textsc{Evaluate}(M_{\Theta'}, \mathcal{D}_t)$ \Comment{accuracy and error set}
\If{$A < \tau$}
  \State $\mathcal{P}_t \gets \textsc{BuildPairs}(E)$
  \State $\Theta_{t+1} \gets \textsc{DPO}(\Theta', \mathcal{P}_t)$
\Else
  \State $\Theta_{t+1} \gets \Theta'$
\EndIf
\State $H \gets H \cup \{\text{artifacts, metrics}\}$
\State \Return $\Theta_{t+1}, H$
\end{algorithmic}
\end{algorithm}

\section{Conclusion}
We presented ALAS, a practical system for autonomous model updating that converts web research into high quality training data and integrates it into a base model via SFT and targeted preference optimization. The system is intentionally modular and reproducible: every stage has a clear contract, intermediates are persisted, and runs can be resumed from checkpoints. Empirically, on fast moving domains such as Python releases and security CVEs, ALAS raises post-cutoff accuracy from low baselines (near 0--15\%) to strong performance (about 85--90\%) with one or two iterations, without manual dataset curation.

Relative to retrieval-augmented methods, ALAS trades external memory for persistent parametric knowledge: answers become instantaneous and available offline, while still remaining compatible with retrieval if extra context is desirable. Compared to naive continual pretraining, ALAS focuses training on carefully synthesized Q\&A with provenance, uses evaluation-driven remediation, and applies DPO to correct residual mistakes with minimal additional cost.

From a systems perspective, several design choices were important:
\begin{itemize}
  \item a disciplined artifact schema (curricula, Q\&A, evaluations, manifests) that enables idempotent resumes and audits;
  \item asynchronous generation with tolerant parsing and validation to turn open web content into clean supervision signals at scale;
  \item promotion gates that only adopt a new checkpoint when improvements are reliable and do not regress guarded topics;
  \item a tight evaluation loop with an LLM judge that produces actionable diagnostics to steer curriculum revision.
\end{itemize}

There remain open challenges. Source quality and consensus checking are critical to avoid encoding misinformation. Cost can be significant for very large models, motivating parameter efficient tuning, curriculum optimization, and smaller student models distilled from the updated teacher. Long horizon operation introduces the risk of forgetting; scheduling rehearsal and maintaining topic level memory can mitigate this. Finally, robust generalization should be measured on independently authored exams, not only train style probes.

We see ALAS as a template rather than a single fixed pipeline. In enterprise deployments, the retrieval module can point to private corpora, the trainer can switch to open source stacks, and the judge can be replaced or augmented with human review for high stakes settings. For research, ALAS provides a controllable environment to study questions such as: How much web supervision is needed to internalize a fact family? What combinations of SFT and DPO minimize cost for a target gain? How should an agent allocate a fixed budget across planning, research, and training to maximize improvement per token?

Looking forward, we plan to: (i) add parameter efficient adapters so that updates are small and composable, (ii) incorporate consensus and fact checking before data admission, (iii) schedule rehearsal across sessions to reduce forgetting, (iv) instrument the evaluation loop with harder, independently generated exams, and (v) support online triggering based on change detection feeds so that updates happen as soon as relevant information appears. Together, these directions move toward trustworthy, economical, and always current language models that can be maintained by autonomous learning agents.

\section*{Acknowledgments}
We acknowledge open-source tooling and documentation that supported this work, including OpenAI platform resources and the LangGraph ecosystem.

\bibliographystyle{plainnat}
\bibliography{references}

\begin{thebibliography}{15}
\providecommand{\natexlab}[1]{#1}
\providecommand{\url}[1]{\texttt{#1}}
\expandafter\ifx\csname urlstyle\endcsname\relax
  \providecommand{\doi}[1]{doi: #1}\else
  \providecommand{\doi}{doi: \begingroup \urlstyle{rm}\Url}\fi

\bibitem[lan()]{langgraph_docs}
Langgraph documentation.
\newblock \url{https://langchain-ai.github.io/langgraph/}.
\newblock Accessed 2025-08-13.

\bibitem[ope()]{openai_platform}
Openai platform documentation.
\newblock \url{https://platform.openai.com/docs}.
\newblock Accessed 2025-08-13.

\bibitem[aut(2023)]{autogpt2023}
Autogpt.
\newblock \url{https://github.com/Significant-Gravitas/AutoGPT}, 2023.
\newblock Accessed 2025-08-13.

\bibitem[bab(2023)]{babyagi2023}
Babyagi.
\newblock \url{https://github.com/yoheinakajima/babyagi}, 2023.
\newblock Accessed 2025-08-13.

\bibitem[Gururangan et~al.(2020)Gururangan, Marasovi{\'c}, Swayamdipta, Lo,
  Beltagy, Downey, and Smith]{gururangan2020dontstop}
Suchin Gururangan, Ana Marasovi{\'c}, Swabha Swayamdipta, Kyle Lo, Iz~Beltagy,
  Doug Downey, and Noah~A. Smith.
\newblock Don't stop pretraining: Adapt language models to domains and tasks.
\newblock In \emph{Proceedings of the 58th Annual Meeting of the Association
  for Computational Linguistics}, 2020.

\bibitem[Guu et~al.(2020)Guu, Lee, Tung, Pasupat, and Chang]{guu2020realm}
Kelvin Guu, Kenton Lee, Zora Tung, Panupong Pasupat, and Ming-Wei Chang.
\newblock {REALM}: Retrieval-augmented language model pre-training.
\newblock In \emph{Proceedings of the 37th International Conference on Machine
  Learning (ICML)}, 2020.
\newblock URL \url{https://arxiv.org/abs/2002.08909}.

\bibitem[Izacard and Grave(2021)]{izacard2021fid}
Gautier Izacard and Edouard Grave.
\newblock Leveraging passage retrieval with generative models for open domain
  question answering.
\newblock \emph{arXiv preprint arXiv:2007.01282}, 2021.
\newblock URL \url{https://arxiv.org/abs/2007.01282}.

\bibitem[Karpukhin et~al.(2020)Karpukhin, Oğuz, Min, Wu, Edunov, Chen, and
  Yih]{karpukhin2020dpr}
Vladimir Karpukhin, Barlas Oğuz, Sewon Min, Ledell Wu, Sergey Edunov, Danqi
  Chen, and Wen-tau Yih.
\newblock Dense passage retrieval for open-domain question answering.
\newblock In \emph{Proceedings of the 2020 Conference on Empirical Methods in
  Natural Language Processing (EMNLP)}, 2020.
\newblock URL \url{https://arxiv.org/abs/2004.04906}.

\bibitem[Lewis et~al.(2020)Lewis, Perez, Piktus, Karpukhin, Goyal, Kukla, Fan,
  Lewis, Yih, Rockt{"a}schel, et~al.]{lewis2020retrieval}
Patrick Lewis, Ethan Perez, Aleksandra Piktus, Vladimir Karpukhin, Naman Goyal,
  Heinrich Kukla, Angela Fan, Mike Lewis, Wen-tau Yih, Tim Rockt{"a}schel,
  et~al.
\newblock Retrieval-augmented generation for knowledge-intensive nlp.
\newblock In \emph{Advances in Neural Information Processing Systems}, 2020.

\bibitem[Meng et~al.(2022)Meng, Bau, Andonian, and Belinkov]{meng2022rome}
Kevin Meng, David Bau, Alex Andonian, and Yonatan Belinkov.
\newblock Locating and editing factual associations in {GPT}.
\newblock In \emph{Advances in Neural Information Processing Systems}, 2022.
\newblock URL \url{https://arxiv.org/abs/2202.05262}.

\bibitem[Meng et~al.(2023)Meng, Andonian, Belinkov, and Bau]{meng2023memit}
Kevin Meng, Alex Andonian, Yonatan Belinkov, and David Bau.
\newblock Mass-editing memory in a transformer.
\newblock \emph{arXiv preprint arXiv:2210.07229}, 2023.
\newblock URL \url{https://arxiv.org/abs/2210.07229}.

\bibitem[Rafailov et~al.(2023)Rafailov, Sharma, Mitchell, Ermon, and
  Finn]{rafailov2023dpo}
Rafael Rafailov, Archit Sharma, Eric Mitchell, Stefano Ermon, and Chelsea Finn.
\newblock Direct preference optimization: Your language model is secretly a
  reward model.
\newblock \emph{arXiv preprint arXiv:2305.18290}, 2023.
\newblock URL \url{https://arxiv.org/abs/2305.18290}.

\bibitem[Shinn et~al.(2023)Shinn, Labash, Veer, Hausknecht, Liu,
  et~al.]{shinn2023selfrefine}
Noah Shinn, Federico Labash, Ashwin Veer, Matthew Hausknecht, Weizhu Liu,
  et~al.
\newblock Self-refine: Iterative refinement with self-feedback.
\newblock \emph{arXiv preprint arXiv:2303.17651}, 2023.
\newblock URL \url{https://arxiv.org/abs/2303.17651}.

\bibitem[Zheng et~al.(2023)Zheng, Chiang, Sheng, Zhang, Li, Li, Li, Li, Sun,
  Du, et~al.]{zheng2023judge}
Lianmin Zheng, Wei-Lin Chiang, Ying Sheng, Siyuan Zhang, Yong Li, Zi~Li,
  Zhanghao Li, Eric Li, Tianjun Sun, Nan Du, et~al.
\newblock Judging {LLM}-as-a-{Judge}: Benchmarking {LLM}s as automatic judges.
\newblock \emph{arXiv preprint arXiv:2306.05685}, 2023.
\newblock URL \url{https://arxiv.org/abs/2306.05685}.

\bibitem[Zweiger et~al.(2025)]{seal2025}
N.~Zweiger et~al.
\newblock Self-adapting language models.
\newblock \emph{arXiv preprint arXiv:2506.10943}, 2025.
\newblock URL \url{https://arxiv.org/abs/2506.10943}.

\end{thebibliography}

\end{document}